# TEMPORAL-CHANNEL TOPOLOGY ENHANCED NETWORK FOR SKELETON-BASED ACTION RECOGNITION

*Jinzhao Luo, Lu Zhou, Guibo Zhu, Guojing Ge, Beiying Yang, Jinqiao Wang*

National Laboratory of Pattern Recognition, Institute of Automation, Chinese Academy of Sciences
School of Artificial Intelligence, University of Chinese Academy of Sciences
luojinzhao2020@ia.ac.cn, {lu.zhou, gbzhu, guojing.ge, beiying.yang, jqwang}@nlpr.ia.ac.cn

## ABSTRACT

Skeleton-based action recognition has become popular in recent years due to its efficiency and robustness. Most current methods adopt graph convolutional network (GCN) for topology modeling, but GCN-based methods are limited in long-distance correlation modeling and generalizability. In contrast, the potential of convolutional neural network (CNN) for topology modeling has not been fully explored. In this paper, we propose a novel CNN architecture, Temporal-Channel Topology Enhanced Network (TCTE-Net), to learn spatial and temporal topologies for skeleton-based action recognition. The TCTE-Net consists of two modules: the Temporal-Channel Focus module, which learns a temporal-channel focus matrix to identify the most critical feature representations, and the Dynamic Channel Topology Attention module, which dynamically learns spatial topological features, and fuses them with an attention mechanism to model long-distance channel-wise topology. We conduct experiments on NTU RGB+D, NTU RGB+D 120, and FineGym datasets. TCTE-Net shows state-of-the-art performance compared to CNN-based methods and achieves superior performance compared to GCN-based methods. The code is available at https://github.com/aikuniverse/TCTE-Net.

## 1. INTRODUCTION

Action recognition is a crucial task with applications in various fields such as human-robot interaction and virtual reality. With the continuous development of depth sensors and pose estimators, obtaining high-quality 3D skeletal data has become easier. As a result, skeleton-based action recognition received increasing attention in recent years, thanks to the compactness and robustness of human skeletal data against complicated backgrounds.

Graph Convolutional Networks (GCNs) [1, 2, 3] have become one of the most popular skeleton-based action recognition methods due to its ability to handle irregular topological information in skeletons [4, 5]. Specifically, GCNs model skeleton sequences as spatiotemporal graph topologies. ST-GCN [5], the first well-known GCN-based method, constructs spatial and temporal correlations in skeletal data via graph convolution. Subsequently, Li *et al*. [6] expands the receptive field based on the self-attention mechanisms to learn topology between joints, while Wang *et al*. [7] aggregates the spatiotemporal topological feature representations to improve the modeling capacity. However, GCN-based methods have limitations. Joint nodes in skeleton are treated equally, which means important nodes and edges cannot be identified [20]. Furthermore, GCNs struggle to model the complicated correlations between distant unnaturally connected joint nodes. Besides, GCN-based methods require complex network structure designs to fuse skeleton and other modalities [8].

Compared with GCN, CNN can model topological features more effectively with powerful local convolution characteristics and self-attention mechanism [9, 10], They can also be easily fused with other modalities [11]. Caetano *et al*. [12] converts the skeleton coordinates to a three channels pseudo image input, and then classifies the features extracted through the network. Such input cannot exploit the locality nature of convolution networks. PoseC3D [11] generates 3D heatmap volumes from skeleton coordinates as input, and then classifies with a 3D-CNN. However, existing CNN-based methods do not utilize the natural topology of the bones. In order to solve the above problems, we propose a novel Temporal-Channel Topology Enhanced Network (TCTE-Net), which models the topological information of skeleton data effectively. Specifically, we propose a novel Temporal-Channel Focus (TCF) module and a Dynamic Channel Topology Attention (DCTA) module, which are used to emphasize critical features and model the correlation between distant joint nodes dynamically.

Our contributions are summarized as follows:

1. We propose TCTE-Net for skeleton-based action recognition equipping with TCF and DCTA modules. The TCF module emphasizes the critical joint nodes with a focus matrix. DCTA learns distant channel-wise topology modeling based on the dynamic channel distance matrix and attention mechanism.

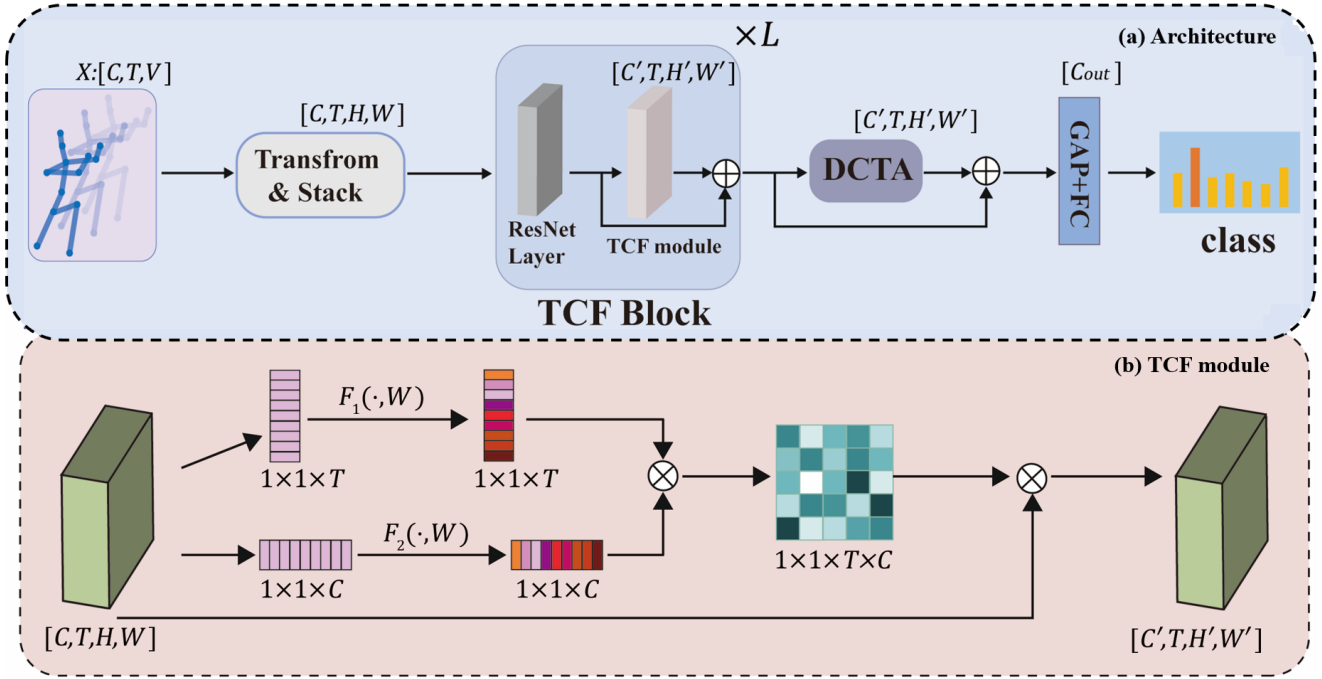

**Fig. 1**: (a) Pipeline of the proposed TCTE-Net, which consists of L TCF blocks and one DCTA module. We instantiate TCTE-Net with the SlowOnly backbone, where L is 3. (b) The detailed architecture of TCF module.

2. The extensive experiments verify the effectiveness of TCF and DCTA modules. The proposed TCTE-Net outperforms state-of-the-art CNN methods significantly and achieves remarkable performance compared to GCN-based methods on three skeleton-based action recognition datasets.

## 2. PROPOSED METHOD

Joint nodes in different body parts contribute to action classification differently. For example, in the case of a 'shaking hands' action, the weight of arm part is much higher than the body part. However, existing CNN-based methods are limited in identifying important joint nodes, and CNNs cannot model the natural topology of the bones directly without an adjacency matrix, which is widely used in GCNs. To address these limitations, we propose the Temporal-Channel Focus module and the Dynamic Channel Topology Attention module, which is introduced in details in the following sections.

### 2.1. Network Architecture

The TCTE-Net is illustrated in Fig. 1(a), which consists of three TCF blocks and one DCTA module. TCTE-Net adopts lightweight SlowOnly 3D-CNN [11, 13] as the backbone. Our approach focuses more on feature representation and topology modeling. Specifically, a 2D skeleton coordinate is represented as a heatmap of size $K \times H \times W$, where $K$ is the number of joints, $H$ and $W$ are the height and width of the frame. All heatmaps are stacked along the temporal dimension $T$. We adopt the 3D heatmap volumes as input. The input joints dimension can be viewed as image channels dimension. In this case, a joint node in the original 2D skeleton is represented as a heatmap of size $H \times W$. The 3D heatmap volumes are fed directly into the network directly and convert into high-level features through ResNet backbones. Next, we extract critical joints feature representations through TCF modules and model the topological relationship between distant joint nodes through the DCTA module. Finally, a classifier is followed to predict action labels.

### 2.2. Temporal-Channel Focus Module

To enhance the important joint nodes, we propose TCF module, which is illustrated in Fig. 1(b). Before being fed into TCF module, the input features are transformed into high-level representations $X \in \mathbb{R}^{C \times T \times H \times W}$. Global average pooling (GAP) is implemented on $X$ along the temporal dimension and channel dimension, followed with a FC layer respectively. Through the above operations, we get the weighted vectors of joint node features in spatial and temporal dimensions. The weighted vectors are then fused by element-wise multiplication. Activation function is applied to get the temporal-channel focus matrix. In the early stages of the network, the heatmaps of different channels represent skeleton joint features of different body parts, which has different weights in classification. Therefore, our TCF module effectively enhances the critical joint nodes in classification based on the weighted temporal-channel focus matrix.

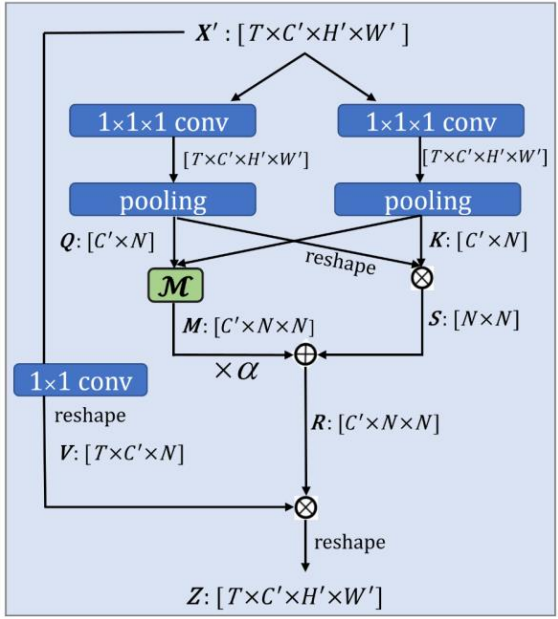

**Fig. 2**: The detailed architecture of DCTA module.

Finally, the joint features are strengthened by temporal-channel focus matrix. The overall process of TCF can be formulated as:

$$X' = \sigma(F_1(GAP(X)) * F_2(GAP(X))) * X ,\quad (1)$$

where $X \in \mathbb{R}^{C \times T \times H \times W}$ and $X' \in \mathbb{R}^{C' \times T \times H' \times W'}$ are the input and output of the TCF module respectively. $F_1(\cdot)$ and $F_2(\cdot)$ mean FC layers. $\sigma(\cdot)$ is activation function.

### 2.3. Dynamic Channel Topology Attention Module

To eliminate the weakness of CNN in modeling the irregular skeletal topology, we introduce the Dynamic Channel Topology Attention module (DCTA). As shown in Fig. 2, the self-attention matrix is used to extract global shared topology for all channels. Meanwhile, we learn specific relationships between joint nodes of different channels.

We utilize convolution and pooling operations on input feature $X'$ to generate new feature representations $Q, K, V$, and reshape to $Q, K \in \mathbb{R}^{C' \times N}$ and $V \in \mathbb{R}^{C' \times T \times N}$, where $N = H' \times W'$. Then we calculate the spatial attention map $S \in \mathbb{R}^{N \times N}$:

$$S_{ji} = \frac{\exp(q_i \cdot k_j)}{\sum_{i=1}^{N} \exp(q_i \cdot k_j)} ,\quad (2)$$

where $S_{ji}$ measures the correlation between position $q_i$ and $k_j$. CNN is capable of modeling the topology of joints implicitly [14]. Thus $S$ are adopted to represent the global shared topological relationship between features of different joint nodes. Meanwhile, we calculate the channel-specific correlations between different features $M \in \mathbb{R}^{C' \times N \times N}$, which can be formulated as:

$$M(x_i, x_j) = \sigma(Q(x_i) - K(x_j)) ,\quad (3)$$

where $\sigma(\cdot)$ is activation function. $M(x_i, x_j)$ is dynamic channel distance matrix, which calculates distances between features $x_i$ and $x_j$ along channel dimension. different channels represent different types of motion features in classification [1]. Therefore, $M$ essentially models the topological relationship between joint nodes under different motion features. For distant unnaturally connected joint nodes, $M$ is able to capture their specific correlations under different motion features dynamically. The final topological relation $R \in \mathbb{R}^{C' \times N \times N}$ is formulated as:

$$R = S + \alpha \cdot M .\quad (4)$$

The dynamic channel distance matrix $M$ is utilized to enhance the global shared topological representation $S$ with a trainable scalar $\alpha$. The addition is conducted in a broadcast way. Finally, we perform a matrix multiplication between $R$ and $V$, and reshape the result to $\mathbb{R}^{C' \times T \times H' \times W'}$. The output of DCTA module is formulated as:

$$Z_j = X_j' + \sum_{i=1}^{N} r_{ji} v_i .\quad (5)$$

The Equation 5 shows that the output feature $Z$ is the sum of the final topological relation and original features, which models the long-range correlations between joint nodes dynamically.

## 3. EXPERIMENTS

### 3.1. Datasets and Implementation Details

**NTU RGB+D** [15] is a large-scale human action recognition dataset. It contains more than 56K video samples of 60 human action classes performed by 40 distinct human subjects. Each sample is captured from different views by three Microsoft Kinect v2 cameras at the same time. The dataset has two benchmarks: Cross-subject (X-Sub), Cross-view (X-View), for which are split by action subjects, camera views in training and validation.

**NTU RGB+D 120** [16] extends NTU RGB+D with 57k video samples of additional 60 action classes, which contains 113k samples over 120 human action classes performed by 106 human subjects. The authors recommend two benchmarks: Cross-subject (X-Sub) and Cross-setup (X-Set, split by camera setups).

**Table 1**: Ablation study on NTU RGB+D. FM represents Focus Matrix.

| Method | Param. | X-Sub | X-View |
|---|---|---|---|
| Baseline | 2.03M | 93.3 | 96.2 |
| +TCF (3) | +0.25M | 93.7 | 96.5 |
| +TCF (3) w/o FM | +0.25M | 93.5 | 96.4 |
| +DCTA | +0.30M | 93.6 | 96.5 |
| + TCF (3) + DCTA | +0.53M | **93.8** | **96.6** |

**Table 2**: Comparison of the classification accuracy (%) with the state-of-the-art methods on the NTU RGB+D dataset. + denotes the result of four-stream fusion.

| Type | Method | X-Sub | X-View |
|---|---|---|---|
| CNN | DSTA-Net [18] | 91.5 | 96.4 |
| | Ta-CNN+ [14] | 90.7 | 95.1 |
| | PoseConv3D [11] | 93.7 | 96.6 |
| GCN | MS-G3D+ [3] | 91.5 | 96.2 |
| | STF [19] | 92.5 | 96.9 |
| | HD-GCN+ [20] | 93.0 | **97.0** |
| Ours | TCTE-Net | **93.8** | 96.6 |

**FineGYM** [17] is a fine-grained action recognition dataset. It contains 29K videos of 99 fine-grained action classes collected from 300 professional gymnastics competitions.

**Implementation details.** TCTE-Net is implemented via Pytorch and trained with 8 RTX 2080 TI GPUs, where each GPU has 11 video clips in a mini-batch. The model is trained for 30 epochs with SGD optimizer. The initial learning rate is set to 0.1375 and decayed with Cosine Annealing scheduler [21]. The weight decay is set to 0.0003. For all datasets, we report the results of 10-clip testing.

### 3.2. Ablation Study

**Effectiveness of TCF and DCTA.** Table 1 illustrates the performance gains brought about by TCF and DCTA on the NTU RGB+D dataset. We use SlowOnly as the baseline model and add the TCF module to it. Our results show that TCF boosts accuracy by 0.4% on X-view benchmark, with little parameter increase, thus validating the effectiveness of TCF. We then remove the Focus Matrix (FM) from TCF to evaluate its impact. The TCF without FM module parallels the weighted channel and temporal vectors. Compared to TCF, the performance of TCF without FM drop by 0.2%, indicating that the weighted focus matrix can efficiently enhance the joint features. Finally, by introducing the DCTA module, we further improve accuracy by 0.3%. Our proposed TCTE-Net achieves an accuracy of 93.8% with the X-Sub benchmark, which improves the baseline accuracy by 0.5% with an efficient model.

### 3.3. Comparison with the State-of-the-Art

In the experimental results section, we evaluate the effectiveness of TCTE-Net on three benchmark datasets: NTU RGB+D, NTURGB+D 120, and FineGYM. Many state-of-the-art methods employ multi-stream fusion models [20, 22, 23], *i.e.*, joint, bone. For a fair comparison, we compare our model with the state-of-the-art methods obtained by the best single models on each dataset, and our model significantly outperforms the other methods.

On the NTU RGB+D dataset, the results shown in Table 2 demonstrate that our model is effective. Although the performance of X-View benchmark is nearly saturated, our model still obtains remarkable performance. Moreover, we achieve an accuracy of 93.8% for X-Sub benchmarks, which outperforms other state-of-the-art methods.

**Table 3**: Comparison of the classification accuracy (%) with the state-of-the-art methods on the NTU RGB+D 120 dataset.

| Type | Method | X-Sub | X-Set |
|---|---|---|---|
| CNN | DSTA-Net [18] | 86.6 | 89.0 |
| | Ta-CNN+ [14] | 85.7 | 87.3 |
| | PoseConv3D [11] | 86.0 | 89.6 |
| GCN | Shift-GCN [24] | 85.9 | 87.6 |
| | InfoGCN [25] | 85.1 | 86.3 |
| | HD-GCN [20] | 85.7 | 87.3 |
| Ours | TCTE-Net | **86.6** | **89.9** |

**Table 4**: Comparison of the top-1 accuracy with the state-of-the-art methods on FineGYM.

| Method | Mean Top-1 Accuracy (%) |
|---|---|
| MS-G3D+ [3] | 92.6 |
| PoseConv3D [11] | 93.2 |
| TCTE-Net | **93.8** |

On the challenging NTU RGB+D 120 dataset, our model achieves excellent performance, as shown in Table 3. We obtain 0.3% improvements for X-Set benchmarks compared with the state-of-the-arts.

Furthermore, we evaluate TCTE-Net on the FineGYM dataset. The results shown in Table 4 demonstrate that our model achieves state-of-the-art performance on the FineGYM dataset. Our model obtains an accuracy of 93.8%, which outperforms the state-of-the-art GCN-based method by 1.2%. Notably, the GCN-based methods are weak in modeling non-connected joint relationships, while our model is able to capture long-range correlations of non-directly connected joints in the skeleton. Therefore, for FineGYM with large movement and deformation, TCTE-Net achieves higher performance than the GCN-based methods.

## 4. CONCLUSION

This paper proposes TCTE-Net, a novel framework for skeleton-based action recognition that addresses the limitations of CNN in modeling the irregular topology of the skeletal data. Through the proposed Temporal-Channel Focus module and Dynamic Channel Topology Attention module, we enhance the ability of TCTE-Net to identify critical joint nodes and model the correlation between joints under different motion features. Experiments on three benchmark datasets show that TCTE-Net outperforms the previous state-of-the-art models. Our work contributes to exploring the potential of CNNs for modeling skeletal data, and we hope that this will inspire further investigations in this direction.